# Show me your NFT and I tell you how it will perform: Multimodal representation learning for NFT selling price prediction[*]

Davide Costa
davide.costa@dimes.unical.it
DIMES, University of Calabria
Rende (CS), Italy

Lucio La Cava
lucio.lacava@dimes.unical.it
DIMES, University of Calabria
Rende (CS), Italy

Andrea Tagarelli
andrea.tagarelli@unical.it
DIMES, University of Calabria
Rende (CS), Italy

## ABSTRACT

Non-Fungible Tokens (NFTs) represent deeds of ownership, based on blockchain technologies and smart contracts, of unique crypto assets on digital art forms (e.g., artworks or collectibles). In the spotlight after skyrocketing in 2021, NFTs have attracted the attention of crypto enthusiasts and investors intent on placing promising investments in this profitable market. However, the NFT financial performance prediction has not been widely explored to date.

In this work, we address the above problem based on the hypothesis that NFT images and their textual descriptions are essential proxies to predict the NFT selling prices. To this purpose, we propose MERLIN, a novel multimodal deep learning framework designed to train Transformer-based language and visual models, along with graph neural network models, on collections of NFTs' images and texts. A key aspect in MERLIN is its independence on financial features, as it exploits only the primary data a user interested in NFT trading would like to deal with, i.e., NFT images and textual descriptions. By learning dense representations of such data, a price-category classification task is performed by MERLIN models, which can also be tuned according to user preferences in the inference phase to mimic different risk-return investment profiles. Experimental evaluation on a publicly available dataset has shown that MERLIN models achieve significant performances according to several financial assessment criteria, fostering profitable investments, and also beating baseline machine-learning classifiers based on financial features.

## 1 INTRODUCTION

*Non-Fungible Tokens*, commonly shortened as *NFTs*, represent today one of the most fashionable applications of the blockchain technology, as well as the trailblazers for the advent of the *Web3*.

Governed by smart contracts, i.e., clauses encoded in programming language that can be deployed using cryptographically signed transactions on the blockchain, NFTs represent pieces of information stored on the blockchain that certify the uniqueness of digital assets. And being traceable, NFTs enable inspecting the history of a given asset, from its creation (or *minting*) to all its owners, up to the present.

The potential shown by NFTs immediately attracted many digital creators and corporations, interested in certifying assets of various types, from images to in-game objects, video and audio contents. All of that suddenly translated in a fervor of investments that peaked more than $2 billion USD traded in the first quarter of 2021,[1] making NFTs a global-scale phenomenon. Indeed, such an enthusiasm led to events never seen before in the art landscape,[2] such as cute kittens (i.e., *CryptoKitties*) capable of congesting the Ethereum network,[3] the third highest price of a living artist in an auction (i.e., Beeple's Everydays: The First 5000 Days, for $69.3 Million),[4] automatically generated pixel artworks (i.e., *CryptoPunks*) selling for nearly $24 Million,[5] or the sale of the first Tweet for over $2.9 Million.[6]

To date, NFTs appear to lead the development of flourishing applications in the increasingly expanding Web3: whether they are deeds of ownerships in the *Metaverse* or tokens aimed at the preservation of cultural heritage,[7] NFTs are lending themselves as representatives of the new era of the Web. In this respect, there is a demand for tools that can unveil their potential and predict their financial performance, thus fostering novel investments and a better growth of the domain.

**Related work.** The skyrocketing of NFTs in 2021 attracted the attention of numerous research groups intent on unraveling opportunities and challenges manifested by an infant yet impactful technology in the blockchain domain [6, 7, 9, 10, 12, 13, 17–19, 22, 23], which served as a forerunner for the Web3.

In [23], Wang et al. describe technical components, protocols, standards, and desired properties for the state-of-the-art NFT solutions. The seminal work by Nadini et al. [13] represents to date one of the most interesting research contributions on NFTs. Based on about 6M transactions concerning nearly 5M NFTs collected between 2017 and 2021 from the Ethereum and WAX blockchains, Nadini et al. unveiled that most traders specialize in particular collections, NFTs in a collection tend to be visually homogeneous, and their visual features can improve the price predictability of NFTs compared to using the transactions history alone. Vasan et al. [19] studied the Foundation platform from a network analysis perspective, based on more than 48k NFTs listed by over 15k artists. In that study, the order of landing on the platforms of artists and investors was found to affect the chances of earning and spending, respectively. Also, in contrast to what is commonly observed for "traditional" art, fluctuations in asset prices are detected for the same creator; nonetheless, these are found to be in a stable range, which also determines the creator reputation. In addition, ties between artists and collectors are crucial to developing a dense network of investment that endures over time.

---

[1]https://nonfungible.com/reports/2021/en/q1-quarterly-nft-market-report
[2]https://decrypt.co/62898/most-expensive-nfts-ever-sold
[3]https://qz.com/1145833/cryptokitties-is-causing-ethereum-network-congestion/
[4]https://www.christies.com/about-us/press-archive/details?PressReleaseID=9970
[5]https://decrypt.co/92819/cryptopunks-ethereum-nft-sells-for-nearly-24m-doubling-previous-record
[6]https://www.cnbc.com/2021/03/22/jack-dorsey-sells-his-first-tweet-ever-as-an-nft-for-over-2point9-million.html
[7]https://cointelegraph.com/news/the-world-s-cultural-heritage-is-being-preserved-one-nft-at-a-time

More recently developed studies have focused on a major challenge in the (crypto) economy, i.e., the ability to model and predict an asset's financial performance, bringing this to the realm of NFTs. Dowling [7] investigated the lack of correlation between the pricing of NFTs and the performance of cryptocurrencies, although the performance of cryptos can actually provide insights into the pricing of NFTs. The influence of intrinsic factors on the price of NFTs was evaluated by Mekacher et al. [12] based on 3.7M transactions collected between 2018 and 2022 for 1.4M NFTs and more than 400 collections. In particular, Mekacher et al. studied the impact of the rarity of (possibly collection-specific) visual attributes (e.g., background, eyes, color) on the financial performance of NFTs, showing that collections are quite heterogeneous in terms of rarity, while rarer NFTs manage to sell at higher prices, with lower negative return risks. By focusing on play-to-earn gaming NFTs dubbed *Axies*, Ho et al. [9] unveiled that intrinsic utility-based features (e.g., attack/defense scores of specific traits) can act as predictors and improve price predictions based just on rarity. Conversely, Kapoor et al. [10] considered external influencing factors, particularly the social influence on the financial performance of NFTs. Through the creation of a dataset linking OpenSea NFTs and corresponding Twitter data, the relationship between asset price and social phenomena was analyzed according to an NFT performance prediction task integrating native (i.e., OpenSea) and exogenous (i.e., Twitter) information. The best-performing model was achieved by exploiting information from both platforms; also, integrating social information showed to improve accuracy by 6% compared to baseline models using NFT-related features only.

It should however be noted that, although [9, 12], resp. [10], are the first to deal with intrinsic, resp. extrinsic, factors on the NFT pricing, the visual/utility features exploited by the former approaches are not learned and correspond to simple descriptive metadata stored in publicly available platforms;[8,9] moreover, the latter approach is limited to the OpenSea market and relies on pseudo-financial features, along with information from social media, to address the prediction task according to hand-crafted classes. Also, there exist tools, like SwapSpace,[10] that offer price projections from various forecast platforms, but they are designed to provide predictions in the form of aggregate forecasts only.

Besides the above limitations, we notice that *no work has been proposed so far to predict the financial performance of an NFT based exclusively on its raw data, i.e., just an image and associated textual description*. We believe that this perspective on the NFT performance prediction problem can offer a number of advantages: (i) it corresponds to a more natural statement of the problem which only requires the context of the input data, (ii) it avoids depending on feature engineering tasks, i.e., the selection of financial explanatory variables for the prediction task, and (iii) by discarding any requirements in terms of prior knowledge on the financial domain (i.e., NFT market), it would better support newbies in NFT trading.

**Contributions.** In light of the above remarks, in this work we address the following problem: based on the key assumption that a (possibly non-expert) user interested in trading NFTs looks at their raw contents as primary source to deal with, given an NFT image and associated textual description, we want to predict the NFT financial performance in terms of selling price. To this purpose, we propose a novel bimodal deep-learning framework that is designed to train Transformer-based language models and visual models, along with graph neural network models, on collections of NFT images and texts. The objective is to learn dense representations from the NFT images and texts, upon which a price classification task is carried out. To accomplish this, our approach remains independent from any NFT financial feature engineering task, as it just requires to know the categories of selling prices for the NFTs in input to the training phase. Yet, in the inference phase, our framework can make flexible predictions according to different user's preferences reflecting a risk-return investment principle. By means of the proposed framework, we aim at answering the following research questions:

**(RQ1)** *Catch'em all* — Can we learn a model capable of suggesting profitable investments in NFT trading?

**(RQ2)** *Take it easy* — Can we answer **RQ1** by just looking at intrinsic contents of NFTs, i.e., images and textual descriptions, thus discarding any requirements in terms of financial indicators?

**(RQ3)** *Attention is all you want* — Is it more valuable to be a "good artist" or a "good writer"? i.e., What is the impact on the NFT performance prediction by the image, its textual description, and an attentive combination of both?

**(RQ4)** *Be aware of your neighbors* — What is the effect of introducing contextual awareness learned from similarity-induced aggregate information?

**Scope and limitations.** To the best of our knowledge, this work is the first to propose a deep-learning financial-agnostic solution to the problem of NFT price-category prediction. The expected impact of our study is hence twofold: raising the bar to new horizons of NFT financial performance prediction, and gaining insights into influencing factors for the value of digital assets in the Web3.

It should be noted that our experimental findings derive from a publicly available dataset, which is, to date, the most representative one on NFT transactions. Although it clearly cannot provide a full picture of the NFT trading realm, we nonetheless point out that our results can reasonably be considered generalizable, due to the high coverage and diversification of the data we used, which (i) include NFTs from heterogeneous platforms (i.e., designed for different scopes), and (ii) cover the NFT trading history up to the steady-state of mid-2021, thus allowing us to catch the main consolidated patterns appeared in the market; in this regard, extreme bull-/bear-driven events of 2022 were left out of our evaluation. Also, our findings are not aware of external sources of influence (e.g., from social media) on the NFT market, neither it considers the history of NFT prices to account for trend analysis; nonetheless, while such aspects are definitely worthy of investigation in future research, the focus of our study is deliberately on information within everyone's reach, thus not requiring any financial domain experience.

---

[8]https://rarity.tools/
[9]https://nonfungible.com/
[10]https://swapspace.co/price-predictions/nft

## 2 PROBLEM DEFINITION

We formulate the NFT performance prediction problem as follows. Given a collection of $N$ NFT data objects, where each object is a pair consisting of an *image* and a *text*, which corresponds to a description for the image, we assume that each NFT is associated with one or more selling prices, over which we take the mean value. From the distribution of the $N$ NFT average-prices, we derive quantile-based intervals to define a set $C$ of *NFT price categories*.

The NFT performance prediction problem is formulated as a classification problem, where the goal is to predict the price category of a previously unseen NFT. This is accomplished according to learned dense representations of NFT images and texts, which are agnostic of financial indicators on the NFT market. To learn the prediction model, training instances correspond to the individual data objects, with ground-truth label $y_i$ associated to the $i$-th training instance corresponding to the category $C \in C$ that encloses the average-price of the NFT instance. The objective of the task is to minimize the cross-entropy loss function:

$$\mathcal{L} = - \sum_{i=1..N, C \in C} y_{i,C} \ln \hat{y}_{i,C}, \qquad (1)$$

where $y_{i,C}$ is 1 if the $i$-th data object actually belongs to class $C$, 0 otherwise, and $\hat{y}_{i,C}$ is the prediction for the $i$-th data object w.r.t. $C$ from a probability distribution matrix $\widehat{Y} \in \mathbb{R}^{N \times |C|}$.

## 3 THE MERLIN FRAMEWORK

We describe our proposed framework named MERLIN - *Multimodal rEpResentation LearnIng for NFT performance prediction*. We first present design requirements in Section 3.1 and an overview of the framework in Section 3.2. In Sections 3.3 and 3.4, we detail the MERLIN stages and its constituting components.

### 3.1 Design requirements

To pursue the goal of predictive modeling in the NFT financial domain (**RQ1**), our architectural choices are naturally fit by deeply contextualized learning models, as we do not want to rely on (financial) feature engineering tasks (**RQ2**). In this respect, and further motivated by the opportunity of learning dense representations of both images and texts for the task at hand, MERLIN exploits the "de-facto" standard in NLP and computer vision, i.e., *Transformer-based pre-trained models*, as well as in graph representation learning, based on *graph neural network* (GNN) models. Indeed, such architectures match our requirement of avoiding manual or domain-driven selection of prominent features and, by relying on the so-called *attention mechanism* [20], they allow lending more significant weights to features detected as more relevant to the task at hand (**RQ3**).

Transformers were originally defined to model language semantics and non-linear relationships between terms, similarly to sophisticated recurrent and convolutional neural networks; however, by employing bidirectional self-supervised training and an attention mechanism that learns contextual relations between (sub-)words in a text, Transformers are much more effective in capturing subtle and complex lexical patterns, including the sequential structure and long-term dependencies, thus obtaining the most comprehensive local and global feature representations of a text sequence [4, 11, 15]. Analogous considerations apply to the visual component of our input (the NFT images), for which we resort to Vision Transformers (ViT) that have recently emerged in computer vision, proposed by Google [5] and Facebook [3]. While adapting the input representation approach used in NLP to images (i.e., image patches are treated as (sub)-words), this type of Transformers has also shown to benefit from the availability of benchmarks, such as ImageNet, to incorporate supervised learning during pre-training.

Moreover, MERLIN utilizes a GNN in order to model the node relations through a message passing scheme to learn the neighborhood importance of each node [24, 25]. Unlike random-walk-based approaches [8, 14], which consider only nodes co-occurring in a random walk and optimize the embeddings to encode random walk statistics, GNN carries out an aggregation scheme by which each node iteratively combines the neighbors and its own features to obtain a new representation. In MERLIN, this allows us to further enhance the learned representations of NFTs in terms of contextual awareness based on similarity search (**RQ4**).

### 3.2 Overview

We propose a flexible two-stage pipeline, as shown in Figure 1.

- Stage 1 includes the image and text learning modules, which are aimed to produce dense representations, or embeddings, for the visual and textual components of NFTs, respectively. Note that, being multilayer Transformer encoder models, they have similar architecture (e.g., input representation as a combination of token embeddings, position embeddings, segment embeddings), although they might use different pretraining objectives. As the two learners work on separate modalities, the image and text embeddings are then combined together and fed into an attention module to fuse the visual features and the lexical/semantic features.
- Stage 2 starts from the fused visual/text embeddings to build an NFT-set similarity graph, where each node is a group of NFTs that is cohesive by average-price and each edge expresses similarity according to the visual and text features. By treating an aggregation of the fused visual/text embeddings as initial features of the nodes, a GNN module is in charge of learning NFT-set dense representations. The neighborhood aggregations learned from the GNN act as an underlying similarity search context, whose results are eventually leveraged for the downstream classification task.

Note that MERLIN is designed to provide alternative points of prediction. In particular, the embeddings individually learned from the PLM and the PVM, the embeddings learned from the attention module, and the embeddings learned from the GNN module can be fed to a *softmax* that yields the prediction probabilities $\hat{y}$ in an

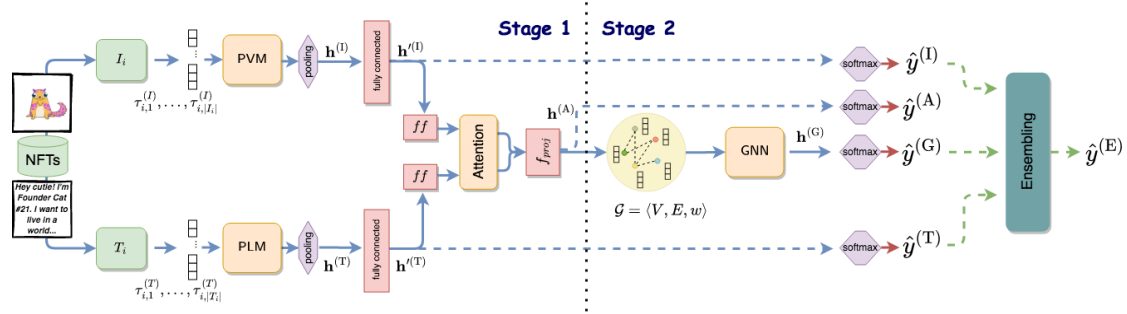

Figure 1: Illustration of the MERLIN framework. Dashed lines refer to pipeline flows alternative to the main flow (solid line)

*alternative* or *complementary* fashion; in the latter case, a multi-view aggregation mechanism (i.e., ensemble) is carried out on all the individual predictions to produce the final prediction.

### 3.3 Stage 1

**Representing input data.** We are given a collection $\mathcal{D}$ of $N$ NFT data, where each data object $D \in \mathcal{D}$ is a pair consisting of an *image*, denoted as $I$, and a *text*, denoted as $T$, which corresponds to a description for the image. Both images and texts are initially represented as token sequences, i.e., for any $D_i = \langle I_i, T_i \rangle \in \mathcal{D}$, $I_i = [\tau_{i,1}^{(I)}, \ldots, \tau_{i,|I_i|}^{(I)}]$ and $T_i = [\tau_{i,1}^{(T)}, \ldots, \tau_{i,|T_i|}^{(T)}]$, where $\tau_{i,j}$ symbols denote the $j$-th tokens of $D_i$ and superscripts (I), (T) are used to denote image and text tokens, respectively.

Clearly, the tokenizers of the two components work on different raw data, and hence the tokens assume different meanings in the two cases: each text is tokenized to yield word pieces (usually to minimize the number of unsegmented words in the text) [4], while each image is split into a sequence of fixed-size non-overlapping patches, which are then linearly embedded [5].

**Learning from texts.** Given an NFT $D_i$, MERLIN employs a Transformer-based pre-trained language model, for short PLM, to deeply contextualize the token sequence of the textual component $T_i$ by mapping it onto a space of dimension $d^{(T)}$, i.e., $\text{PLM}(T_i) \in \mathcal{R}^{|T_i| \times d^{(T)}}$. A pooling function $pooling(\cdot)$ is then applied to the token embeddings to output a single embedding $\mathbf{h}_i^{(T)}$ of size $d^{(T)}$:

$$\mathbf{h}_i^{(T)} = pooling(\text{PLM}(T_i)). \tag{2}$$

Following BERT and related PLMs, a default choice for the pooling function is to output the embedding of the special token [CLS], which is considered to be representative of the whole input text.

**Learning from images.** While working on the image component of NFTs, the image learning module shares with the textual one the architectural choices based on Transformer. Indeed, given a pre-trained visual model, for short PVM, the goal is to deeply encode the token sequence of the visual component $I_i$ into a space of dimension $d^{(I)}$, i.e., $\text{PVM}(I_i) \in \mathcal{R}^{|I_i| \times d^{(I)}}$. Analogously to the text learning component, the token embeddings are subject to a pooling function to yield a single embedding $\mathbf{h}_i^{(I)}$ of size $d^{(I)}$:

$$\mathbf{h}_i^{(I)} = pooling(\text{PVM}(I_i)). \tag{3}$$

Also for PVM, a [CLS] token is added so that its output embedding can be pooled to serve as representation of an entire image.

**Fusing the representations.** To obtain a fused fixed-size dense representation of text and image for each NFT (**RQ3**), we need to jointly encode $\mathbf{h}_i^{(T)}$ and $\mathbf{h}_i^{(I)}$ to a common space. First, we use two fully-connected neural networks to map the two embeddings into a smaller space, $d'^{(T)}$ and $d'^{(I)}$, respectively. On the resulting embeddings, $\mathbf{h}_i'^{(T)}$ and $\mathbf{h}_i'^{(I)}$, a *self-attention* mechanism is carried out. A feed-forward neural network $ff$ computes energy scores for $\mathbf{h}_i'^{(T)}$ and $\mathbf{h}_i'^{(I)}$:

$$e_i^{(I)} = ff(\mathbf{h}_i'^{(I)}), \quad e_i^{(T)} = ff(\mathbf{h}_i'^{(T)}), \tag{4}$$

on which a *softmax* operator is applied to obtain a probability distribution over the two attention energies:

$$\alpha_i^{(I)} = \frac{\exp(e_i^{(I)})}{\exp(e_i^{(I)}) + \exp(e_i^{(T)})}, \quad \alpha_i^{(T)} = \frac{\exp(e_i^{(T)})}{\exp(e_i^{(I)}) + \exp(e_i^{(T)})}. \tag{5}$$

Eventually, we scale our latent representations $\mathbf{h}_i'^{(T)}$ and $\mathbf{h}_i'^{(I)}$ by the corresponding energies, to obtain the following contexts:

$$\mathbf{c}_i^{(I)} = \alpha_i^{(I)} \cdot \mathbf{h}_i'^{(I)}, \quad \mathbf{c}_i^{(T)} = \alpha_i^{(T)} \cdot \mathbf{h}_i'^{(T)} \tag{6}$$

These are then provided to another feed-forward neural network $ff_{proj}$, which projects them onto another space of dimension $d^{(A)}$ (where superscript (A) stands for "attention"):

$$\mathbf{h}_i^{(A)} = ff_{proj}\left(\mathbf{c}_i^{(I)} \oplus \mathbf{c}_i^{(T)}\right), \tag{7}$$

where $\oplus$ denotes the concatenation operator.

### 3.4 Stage 2

**Modeling an NFT similarity context.** The second stage of MERLIN is aimed to enhance the contextual awareness among NFTs by learning new representations based on a similarity-search context built upon the current representations. We define an undirected graph $\mathcal{G} = \langle V, E, w \rangle$, where the node-set $V$ is a partition of $\mathcal{D}$ into $n$ cohesive groups, $E$ is the edge-set expressing similarity relations between nodes (i.e., NFT groups) and $w : E \mapsto \mathbb{R}$ is a weighting function to compute node similarity. The node-set is specified by first ordering the $N$ NFTs by average-price, then forming $n$ groups (i.e., nodes) of almost equal size $l = N/n$ such that each node $v \in V$ contains NFTs having average-prices close within a certain interval of varying length. Moreover, each node is associated with a class label that corresponds to the average-price category.

Given $v \in V$ and an integer $k > 0$, an edge is drawn from $v$ to each of its $k$ most similar nodes, with edge weight defined as:

$$w(v, u) = \cos\left(n\text{-}pooling(\{\mathbf{h}_i^{(A)}\}_{D_i \in v}), n\text{-}pooling(\{\mathbf{h}_j^{(A)}\}_{D_j \in u})\right),$$

where $\cos(\cdot)$ denotes the cosine similarity function, and $n\text{-}pooling(\cdot)$ is a pooling function on the set of NFTs belonging to the same node; by default, we define $n\text{-}pooling(\cdot)$ as the average over the NFT embeddings learned at the end of Stage 1.

**Learning from the graph of NFT-set similarities.** The next step of Stage 2 is to provide the graph $\mathcal{G}$ in input to a GNN module. We denote with $\mathbf{X} \in \mathbb{R}^{|V| \times d^{(A)}}$ the initial-feature (or attribute) matrix associated with the nodes in $\mathcal{G}$, i.e., the embeddings produced by the $n\text{-}pooling$ function. The goal is to learn new node-features $\mathbf{h}^{(G)}$ in a latent space of dimension $d^{(G)}$ (where superscript (G) is for "graph") modeling the relations between cohesive groups of NFTs.

A particularly suitable GNN for our setting is the Graph Attention Network (GAT) [2, 21]. Unlike a graph convolutional network (GCN) that assigns predetermined weights to the neighbors of a node, a GAT learns the weights through a self-attention mechanism in order to capture the importance of different neighbors; more precisely, GAT modifies the aggregation process of GCN by learning the strength of the link between neighboring nodes through self-attention [24]. Formally, the importance of a node $v$'s features w.r.t. node $u$ is computed through the attention coefficients as follows:

$$e_{uv} = a(\mathbf{W}\mathbf{x}_u, \mathbf{W}\mathbf{x}_v) \qquad (8)$$

where $\mathbf{W} \in \mathbb{R}^{d^{(G)} \times d^{(A)}}$ is a trainable weight matrix. The attention mechanism is performed by a feed-forward neural network exploiting the *LeakyReLU* non-linearity function:

$$\alpha_{uv} = \frac{\exp(LeakyReLU(e_{uv}))}{\sum_{k \in \mathcal{N}_u} \exp(LeakyReLU(e_{uk}))}, \qquad (9)$$

where $\mathcal{N}_u$ denotes the set of neighbors of node $u$. At each step, each node $u$ updates its hidden state, denoted as $\mathbf{z}_u$, by aggregating the features of its neighbors as follows:

$$\mathbf{z}_u = \sigma\left(\frac{1}{Q}\sum_{q=1}^{Q}\sum_{v \in \mathcal{N}_u} \alpha_{uv,q} \mathbf{W}_q \mathbf{x}_v\right), \qquad (10)$$

where $Q$ denotes the number of independent attention mechanisms (heads) [20], $\alpha_{uv,q}$ and $\mathbf{W}_q$ are the normalized attention coefficients and weight matrix corresponding to the $q$-th head [21], and $\sigma(\cdot)$ is the *ReLU* activation function.

## 4 EXPERIMENTAL METHODOLOGY

**Data.** We resorted to the most-representative publicly available dataset on NFT purchase transactions [13], which includes sales from the *CryptoKitties*, *Gods-Unchained*, *Decentraland*, *OpenSea*, and *Atomic* markets. It contains 6.1M transactions involving 4.7M NFTs, spanning across more than 4k collections, which were grouped by the creators into six main categories (in parenthesis, we report

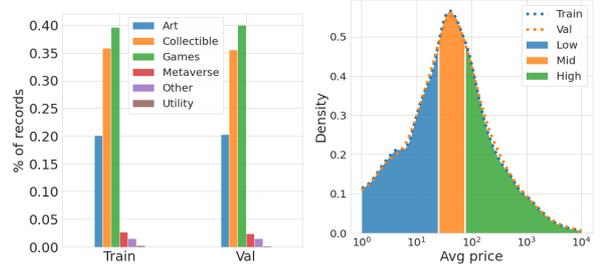

**Figure 2: Histograms of the NFT category distributions for the training and validation sets (left) and density of the average-price distributions (overall, training, and validation sets), with corresponding areas for the three price-categories (i.e., classes) (right)**

the coverage percentage): *Art* (18.46%), *Collectible* (28.85%), *Games* (47.21%), *Metaverse* (0.1%), *Utility* (0.17%), and *Other* (5.21%).

Each NFT is associated with an image, a text description, and the selling prices, which were used to build our training instances. Besides ensuring to filter out NFTs with missing image or text, we also chose to select NFTs having at least a *secondary sell*. This would avoid us incurring such latent patterns as price boosting mechanisms between authors, whose investigation is beyond the objectives of this work (cf. Conclusions). We thus came up with 202,257 NFTs having images (downloaded from the corresponding URLs in the dataset) and descriptions to be used in our experimental evaluation.

As concerns the price categories to be used as class labels for our training data ($C$), we examined the distribution of average-prices and decided to define 3 classes corresponding to the first quarter (i.e., average-prices up to the first quartile), the union of second and third quarter (i.e., average-prices between the first quartile and the third quartile), and the fourth quarter (i.e., average-prices above the third quartile). For short, we hereinafter refer to the 3 classes as *Low*, *Mid*, and *High*, respectively.

We split the dataset into training and validation sets of size 90% and 10%, respectively, by keeping equal per-class distribution w.r.t. the whole dataset, as shown in Figure 2. Also, note that reproducibility of data samples and results is ensured since we fixed the seed-sets for our randomness-related operations.

**Assessment criteria and goals.** To properly evaluate our models so as to quantitatively answer our **RQ1**, we used a number of statistical criteria derived from each confusion matrix outputted by the models (Table 1). These criteria include both global performance measures and more specific performance measures associated with the financial task at hand. The former group contains *accuracy* (*A*), and weighted macro-averaged *precision* (*P*), *recall* (*R*), and *F-score* (*F1*).[11] The latter group focuses on criteria that have clear meanings in terms of trading losses and gains, particularly relating to the most *profitable* (i.e., with a positive-return) investments. This led us to derive the following criteria from the portion of the confusion matrix involving class *High*:

---
[11] We used `scikit-learn` implementations

Davide Costa, Lucio La Cava, and Andrea Tagarelli

Table 1: Confusion matrix for our 3-class classification task (top) and domain-specific criteria (bottom)

|  |  | True Label | | |
|---|---|---|---|---|
|  |  | Low | Mid | High |
| Pred. Label | Low | $C_{L,L}$ | $C_{L,M}$ | $C_{L,H}$ |
|  | Mid | $C_{M,L}$ | $C_{M,M}$ | $C_{M,H}$ |
|  | High | $C_{H,L}$ | $C_{H,M}$ | $C_{H,H}$ |

| criterion | definition | criterion | definition |
|---|---|---|---|
| win rate | $WR = \dfrac{C_{H,H}}{C_{H,L} + C_{H,M} + C_{H,H}}$ (11) | win/loss ratio | $WLR = WR/LR$ (12) |
| loss rate | $LR = \dfrac{C_{H,L} + C_{H,M}}{C_{H,L} + C_{H,M} + C_{H,H}}$ (13) | cautiousness | $Cn = \dfrac{MR}{MR + LR}$ (14) |
| missed opport. rate | $MR = \dfrac{C_{L,H} + C_{M,H}}{C_{L,H} + C_{M,H} + C_{H,H}}$ (15) | riskiness | $Rn = \dfrac{LR}{MR + LR}$ (16) |

- *win rate* (WR), which is defined as in Eq. 11, i.e., the fraction of predicted most-profitable trades that are actually most-profitable (which is also equivalent to the precision for class *High*);
- *win-loss ratio* (WLR), which is defined as in Eq. 12, i.e., the ratio between most-profitable wins and losses;
- *loss rate* (LR), which is defined as in Eq. 13, i.e., the fraction of wrongly predicted most-profitable trades (which is also equivalent to the false discovery rate for class *High*);
- *missed opportunity rate* (MR), which is defined as in Eq. 15, i.e., the fraction of true most-profitable trades that are incorrectly predicted (which is also equivalent to the false negative rate for class *High*);
- *cautiousness* (Cn), which is defined as in Eq. 14, i.e., the ratio of missed opportunities to the total number of errors for class *High*;
- *riskiness* (Rn), which is defined as in Eq. 16, i.e., the ratio of wrong predictions to the total number of errors for class *High* (which is also equivalent to $1 - Cn$).

The above group of statistics, hereinafter referred to as *High-driven performance criteria*, represents a proxy for real-life financial performances. In this regard, note that the win and loss rates are complementary to each other, and determine the win-loss rate, which is strictly related to the *return on investment* (ROI), i.e., the effectiveness of investment choices. Furthermore, we can shape investments on different risk-profiles by looking at the missed opportunity rate; indeed, by identifying "untaken" chances (i.e., $C_{L,H}, C_{M,H}$), it can be used to understand whether the model was risky or cautious on its predictions.

**Models and Settings.** We tested different pre-trained Transformer-based models for both image and text modules. Specifically, we included in our evaluation BERT-base-uncased [4], XML-RoBERTa [11], and S-BERT (all-MiniLM-L6-v2) [15] as PLM, and ViT-base (patch-16-224) [5] and DINO (vitb8) [3] as PVM, which are available in the *HuggingFace* model repository.[12] In all cases, but S-BERT, we set 12 Transformer encoder layers, with hidden size $d^{(I)} = d^{(T)}$ equal to the model default of 768, and 12 attention-heads; as for S-BERT, we used the default 6-layer model. Moreover, we used the [CLS] token embedding as pooling function. Note that PLM and PVM were subject to a fine-tuning stage of training in order to adapt them to our NFT prediction task. Hereinafter, unless otherwise specified, PLM and PVM correspond to BERT and ViT, respectively, which revealed to be our preferred models (cf. Section 5.3).

As concerns the NFT-set similarity graph settings, we varied both its parameters, i.e., the node size $l$ and the node neighborhood size $k$; as we shall discuss later, our best choices for $l$ and $k$ are 50 and 10, respectively. The GNN was implemented through the GATv2Conv [2] module available in PyTorch Geometric,[13] it uses two convolutional layers with 4 concatenated attention heads ($Q$), hidden dimensionality of 16, and dropout probability equal to 0.5. *Please note that investigating the best available PLMs, PVMs, and GNNs for our tasks is beyond the objectives of this work.*

The three linear-projection layers used in MERLIN were equipped with *ReLU* activation function, batch normalization and dropout probability equal to 0.2. The output embedding size was set as $d'^{(I)} = d'^{(T)} = d^{(A)} = 256$.

All models were trained with the *Adam* optimizer and learning rates $1.0E$-5 for both PLM and PVM, and $1.0E$-3 for GAT; the attention module was trained with learning rate $1.0E$-5 when considering PLM and PVM and with $1.0E$-4 with the GAT. We set the number of training epochs to 10, which is relatively large considering that both PLM and PVM are pre-trained models, and that the attention and GAT modules are initialized with contextualized embeddings. Furthermore, early-stopping was applied, saving the model at a maximum validation win-rate, so as to maximize the expected number of profitable predictions. All reported results correspond to averages over ten runs (i.e., different seeds); we noticed very small standard deviation (e.g., order of $1.0E$-4, for accuracy), hinting at high stability of PLM and PVM and, as a consequence due to its weight initialization based on them, of the GAT in cascade.

Our experiments were carried out on a 56-core Intel(R) Xeon(R) Gold 6258R CPU, with 256GB RAM and two NVIDIA GeForce RTX3090s, OS Ubuntu Linux 22.04 LTS.

## 5 RESULTS

### 5.1 Competing baseline methods

We considered two types of baseline methods to be comparatively evaluated w.r.t. our MERLIN models. The first type includes simple yet feature-agnostic models: ZeroR, which always returns the most-frequent price-category as predicted class (i.e., *Mid*), and Prior, which samples the predicted class from the true-class distribution over the input data. The second type of competitors refers to machine-learning classifiers, i.e., SVM and logistic regressor, that are trained over data objects represented by a predetermined set of features. This contains two subsets: the one including min, max, avg and std of the selling prices of the collection an NFT belongs to, and the other one including one-hot encodings of the top-25 most-frequently used terms in the NFT descriptions across $\mathcal{D}$.

As reported in Table 2, ZeroR sets the worst global performance, while being not applicable for the *High*-driven evaluation. The other

---

[12] https://huggingface.co/

[13] https://www.pyg.org/

Multimodal representation learning for NFT selling price prediction feature-agnostic model, Prior, has significantly lower global performance than the machine-learning classifiers, although it behaves comparably to SVM. The logistic regressor is the best competitor according to all criteria but $Rn$ and $Cn$. In any case, however, all baselines are significantly outperformed by our MERLIN models, as we shall describe in the next sections. In particular, it should be noted that all baselines have a $WLR$ much lower than 1, thus being unable to yield profitable predictions.

Table 2: Competing baseline methods

| Baseline | P | R | F1 | A | WR↑ | LR↓ | WLR↑ | MR↓ | Rn↓ | Cn↑ |
|---|---|---|---|---|---|---|---|---|---|---|
| ZeroR | 0.169 | 0.333 | 0.224 | **0.506** | na | na | na | na | na | na |
| Prior | 0.336 | 0.335 | 0.335 | 0.381 | 0.252 | 0.748 | 0.338 | 0.749 | **0.500** | **0.500** |
| SVM | 0.426 | 0.416 | 0.407 | 0.419 | 0.247 | 0.753 | 0.327 | 0.633 | 0.543 | 0.457 |
| Logistic | **0.499** | **0.500** | **0.467** | 0.461 | **0.393** | **0.607** | **0.648** | **0.236** | 0.720 | 0.280 |

Table 3: Summary of results by the best-performing models in MERLIN. First four rows, resp. GAT-based row, refer to training mode, resp. evaluation mode, of the PLM and PVM models. Best values per criterion are in boldface

| Model | Att. | P | R | F1 | A | WR↑ | LR↓ | WLR↑ | MR↓ | Rn↓ | Cn↑ |
|---|---|---|---|---|---|---|---|---|---|---|---|
| PLM | ✗ | 0.767 | 0.738 | 0.720 | 0.738 | 0.891 | 0.109 | 8.136 | 0.622 | 0.150 | 0.850 |
| PVM | ✗ | **0.803** | 0.800 | **0.800** | 0.800 | 0.798 | 0.202 | 3.960 | 0.264 | 0.433 | 0.567 |
| PVM+PLM | ✗ | 0.802 | **0.801** | 0.800 | **0.801** | 0.807 | 0.193 | 4.179 | 0.284 | 0.405 | 0.596 |
| PVM+PLM | ✓ | 0.802 | 0.799 | 0.798 | 0.799 | 0.814 | 0.186 | 4.368 | 0.309 | 0.376 | 0.624 |
| GAT (l=50) on PVM+PLM | ✓ | 0.773 | 0.727 | 0.701 | 0.727 | 0.926 | 0.074 | 12.571 | 0.688 | 0.097 | 0.903 |
| ensemble | - | 0.779 | 0.726 | 0.697 | 0.726 | **0.959** | **0.041** | **23.531** | 0.704 | **0.055** | **0.945** |

### 5.2 Evaluating MERLIN models w.r.t. RQs

Here we discuss the main results achieved by MERLIN models in order to unveil the capability of our proposed framework to answer the RQs stated in the Introduction. Note that the results presented in this section refer to the best performance obtained by the various constituting modules of MERLIN, which are reported in Table 3; sensitivity analysis is postponed to Section 5.3.

**Answering RQ2.** We begin with assessing the individual performances of PLM and PVM. PLM achieves precision, recall, F1, accuracy equal to 0.767, 0.738, 0.720 and 0.738, respectively. Coupling this with the good performance in terms of win-rate (0.891), loss-rate (0.109), and win-loss ratio 8.136, we unveil that the textual descriptions provided by the NFT creators can already serve as valuable information to predict how an NFT will perform. Moreover, the high missed-opportunity rate and the particularly high cautiousness reveal that our fine-tuned PLM tends to be cautious, as it avoids jumping into very risky trading opportunities.

The visual counterpart PVM achieves even higher performance than PLM according to global criteria, which are all not less than 0.800. However, as concerns the $High$-driven performance criteria, we notice a particularly risky model (i.e., minimum $MR$ and $Cn$), which doubles, resp. halves, the loss rate ($LR = 0.202$), resp. win/loss ratio ($WLR = 3.960$), w.r.t. PLM. This might partly be ascribed to the fact that some visual features could not be exclusive of certain collections. As a result, although visual features are certainly valuable to predict the NFT financial performances, they also might lead to a more hazardous and loss-prone behavior.

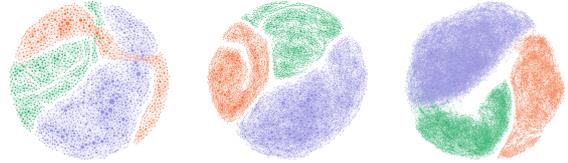

Figure 3: NFT-set similarity graph by varying node size $l$: 100 (left), 50 (center), 25 (right). Colors correspond to the price categories: orange for $Low$, purple for $Mid$, green for $High$. Fruchterman-Reingold layout is used to display the graphs

**Answering RQ3.** To understand the beneficial effects from the combination of both visual and text embeddings, we distinguish two cases, depending on whether the prediction was made on top of the attention module or just on top of the concatenation of the (compressed) visual and text embeddings (i.e., $\mathbf{h}'^{(T)}$ and $\mathbf{h}'^{(I)}$), as reported in the 3rd and 4th rows of Table 3.

While achieving no particular advantages in both cases compared to the best-performing PVM according to global criteria, we notice improvements in terms of the $High$-driven performance criteria, which are particularly evident when the attention module was trained and used for prediction. Overall, the attention-based combination of visual and textual features leads to more cautious predictive behavior than the PVM-only model ($Cn = 0.624$), by fixing some hazardous predictions ($Rn = 0.433$), presumably due to the role of the attention on highly informative textual patterns; also, the attention-based combination has beneficial effects in terms of missed-opportunity rate, which halves w.r.t. the PLM-only model.

**Answering RQ4.** Learning from the similarity relations between price-cohesive groups of NFTs relies on two key parameters, namely the node size $l$ and the node neighborhood size $k$. We focus here on their best-performing settings, whereas we shall discuss the impact due to different settings in Section 5.3. Nonetheless, one important remark that stands out by looking at Figure 3 is that, regardless of a particular setting of $l$, a partition of the graph naturally emerges in terms of the three price-categories. Indeed, as shown in the figure, each of the three classes is well-represented by a densely connected subgraph and, at the same time, the three subgraphs are connected by a few yet non-negligible number of links. This implies that the process of node's neighbor aggregation carried out by the GNN module will involve not only nodes of the same class but also of different classes.

The best-performing GNN results in Table 3 show a striking improvement on financial criteria, by paying just a little in terms of global measures. We notice a remarkably cautious model ($Cn = 0.903$), which peaks up to $WR = 0.926$ and down to $LR = 0.074$, respectively. Accordingly, the score $WLR = 12.571$ is worthy of attention as it increases by 55% versus the best-performing model so far (i.e., the best PLM) with a remarkably better risk-profile.

**Ensembling multi-view predictions.** A further stage of evaluation concerns the opportunity of leveraging the predictions performed by the constituting components of MERLIN, with the goal of providing a more robust prediction according to the $High$-driven criteria. More specifically, rather than aggregating by majority voting or similar criterion, we define a *priority rule*, i.e., a precedence relation among predictors, according to their cautiousness and win-rate: (i) we first consider the most cautious and win-prone model, i.e., GAT, and look at its prediction: if it is not *High*, MERLIN returns the GAT outcome as the predicted class; (ii) if it does not hold, we consider the prediction of the second-most cautious model, i.e., the PLM: again, if it is not *High*, MERLIN returns the PLM outcome as the predicted class; (iii) otherwise, MERLIN entrusts PVM, regardless of the predicted class.

Table 4: Results achieved by MERLIN models with various settings. Symbol ∗ is used to mark the selected model in each subtable. PLM and PVM models are used in training mode, but evaluation mode when supporting the GAT models

|  | Model | Att | P | R | F1 | A | WR ↑ | LR ↓ | WLR ↑ | MR ↓ | Rn ↓ | Cn ↑ |
|---|---|---|---|---|---|---|---|---|---|---|---|---|
| PLM | bert-base-uncased ∗ | ✗ | 0.767 | 0.738 | 0.720 | 0.738 | 0.891 | 0.109 | 8.136 | 0.622 | 0.150 | 0.850 |
|  | xlm-roberta-base-cased | ✗ | 0.769 | 0.738 | 0.719 | 0.738 | 0.900 | 0.100 | 9.010 | 0.628 | 0.137 | 0.863 |
|  | sbert-default | ✗ | 0.740 | 0.719 | 0.700 | 0.719 | 0.831 | 0.169 | 4.909 | 0.641 | 0.209 | 0.791 |
| PVM | vit-base-patch-16-224 ∗ | ✗ | 0.803 | 0.800 | 0.800 | 0.800 | 0.798 | 0.202 | 3.960 | 0.264 | 0.433 | 0.567 |
|  | dino-vit | ✗ | 0.791 | 0.791 | 0.790 | 0.791 | 0.797 | 0.203 | 3.919 | 0.281 | 0.420 | 0.580 |
| PLM+PVM | best PLM, best PVM | ✗ | 0.802 | 0.801 | 0.800 | 0.801 | 0.807 | 0.193 | 4.179 | 0.284 | 0.405 | 0.596 |
|  | best PLM, best PVM ∗ | ✓ | 0.802 | 0.799 | 0.798 | 0.799 | 0.814 | 0.186 | 4.368 | 0.309 | 0.376 | 0.624 |
| GAT | $l$=100, best PLM+PVM | ✓ | 0.771 | 0.731 | 0.708 | 0.731 | 0.914 | 0.090 | 10.560 | 0.660 | 0.115 | 0.885 |
|  | $l$=50, best PLM+PVM ∗ | ✓ | 0.773 | 0.727 | 0.701 | 0.727 | 0.926 | 0.074 | 12.571 | 0.688 | 0.097 | 0.903 |
|  | $l$=25, best PLM+PVM | ✓ | 0.756 | 0.707 | 0.669 | 0.707 | 0.926 | 0.074 | 12.489 | 0.769 | 0.088 | 0.912 |
|  | $l$=50, best PLM+PVM | ✗ | 0.761 | 0.679 | 0.621 | 0.679 | 0.952 | 0.048 | 19.688 | 0.876 | 0.052 | 0.948 |

Looking at Table 3, we find evidence that supports our initial hypothesis: the devised ensemble predictor is particularly cautious ($Cn$ = 0.945) while achieving the best overall results on financial criteria ($WLR$ = 23.531), up to doubling, resp. triplicating, the scores of GAT, resp. PLM; an ever sharper improvement when we consider the PVM, where it results seven times better.

### 5.3 Sensitivity analysis

In this section, we delve into the effect of different choices of the main modules in MERLIN, which complement our discussion in Section 5.2.

Table 4 shows results achieved by various settings for PLM, PVM, and GAT (cf. Section 4). Note that, for PLMs and PVMs, their respective prediction results were obtained at inference time.

We notice that XLM-RoBERTa and BERT behave quite similarly to each other, especially in terms of global criteria, and better than S-BERT. Our choice fell on BERT since, although it achieves slightly lower performance in terms of $WR$ and $LR$, BERT has 40% faster training time w.r.t. RoBERTa according to our early-stopping criterion, while also having less parameters (110M vs. 125M).

As concerns the Transformer-based PVMs, note that they were pre-trained using different approaches: supervised image classification for ViT-base (patch-16-224) [5] and knowledge distillation for DINO (vitb8) [3], albeit they shared the dataset (i.e., ImageNet). The best performer revealed to be ViT, especially due to an increase in performances of 2% w.r.t. DINO.

Regarding the GAT module, the evaluation based on different node-size values (i.e., $l$ = {100, 50, 25}) evidenced roughly comparable performances, with $l$ = 50 and $l$ = 25 ensuring the best global and *High*-driven scores. We narrowed our attention to the setting $l$ = 50 as the best configuration, since it is better in terms of global criteria and slightly improves in terms of $WLR$ w.r.t. the $l$ = 25, while having an inference time noticeably better than the latter (cf. **Appendix A**). We then assessed the impact of the attention component on GAT with $l$ = 50, as shown in the last row of Table 4. In this regard, we noticed that, although the $WLR$ appears to increase, skipping the attention module negatively affects both types of performance criteria, along with a tendency of the resulting model to become too much cautious ($MR$ = 0.876), thus overlooking more promising opportunities.

Moreover, we varied $k$ in {5, 10, 20} while keeping fixed $l$ to 50. Notably, by increasing $k$, $MR$ tends to decrease but with increased $Rn$, which sets on 0.05, 0.097, and 0.75 for $k$ = 5, 10, and 20, respectively. Also, $MR$ = 0.99 obtained for $k$ = 5 hints at a degenerate model.

## 6 LESSONS LEARNED

Here we summarize the results obtained addressing our initially stated research questions, thus providing the reader with a memorandum for the main lessons learned in this work.

- **RQ1**: Deeply contextualized pre-trained learning models undoubtedly represent the most suited choice when we need to address complex tasks in new domains while abstracting from feature engineering. By adapting pre-trained language and visual models to the NFT domain and relating selling-price prediction task, our MERLIN can leverage such models' capabilities to effectively learn meaningful representations of the raw NFT data, i.e., images and their descriptions, to be exploited for financial prediction.
- **RQ2**: MERLIN learned models have shown ability to provide valuable suggestions in the NFT trading landscape without learning from financial features. By contrast, the visual as well as textual features learned by our models from the NFT images and descriptions can serve as helpful financial proxies. Yet, being trained on NFT data which only require knowledge on their average-price category as their class label, our models outperform machine-learning classifiers that were trained over financial features.
- **RQ3**: NFT images and texts convey different yet complementary rich contents for the task at hand. When textual descriptions are really informative and discriminative of NFTs within the same collection, they can suggest profitable investments being fairly cautious w.r.t. risky trading moves. Conversely, visual features might lead to more risky yet still profitable plays in the market, presumably due to their non-uniqueness w.r.t. specific collections and/or creators (e.g., certain visual features are used in multiple collections). Therefore, we would say that being a "good writer" pays more than being a "good artist", unless one wants to try the thrill of risk. Furthermore, an attentive combination of textual and visual embeddings refines the capabilities of MERLIN in a complementary way. Indeed, we spotted that attention shifts towards visual features when the textual component is not informative (e.g., the same description is used for all NFTs in the *Sorare* collection) and, conversely, it leverages the textual component when visual models result excessively risky in predictions.

- **RQ4**: While consistently improving the prediction performances w.r.t. the textual/visual models and their combination, learning from the graph of NFT-set similarities unveils some unexpected yet remarkable patterns. In particular, the embeddings generated by the GAT are highly effective in detecting price categories also at finer resolution (percentiles), and even collections (cf. **Appendix B**) despite never having seen them during the training. It is also worth emphasizing that a multi-view ensemble approach can complement the best skills of the individual modules in MERLIN: a *High*-driven strategy giving precedence among predictors based on a mixture of win-rate and cautiousness, enables a particularly profitable yet reasonably cautious model, which doubles the best individual predictor in terms of financial metrics.

As a final note, remarkable aspects arise from the *explanation* of our MERLIN models. In **Appendix C**, we provide interpretation of the predictions yielded by MERLIN on different examples.

## 7 CONCLUSIONS

We presented MERLIN, a deep-learning-based framework for the task of predicting NFT performance (selling average-price) by solely relying on images and descriptions of NFTs. To the best of our knowledge, this is the first work to address the above task. Code and evaluation data are available at https://people.dimes.unical.it/andreatagarelli/nft-learning/.

We can outline a few of directions for future research. Besides investigating architectural alternatives to the learning modules for further improving the MERLIN performance, it would be interesting to incorporate the temporal dimension (e.g., (re)selling times) into the NFT representation learning. Another line of investigation corresponds to detecting anomalous or adversarial patterns, such as those related to price boosting mechanisms between the NFT creators and traders. Yet, accounting for external sources of influence, such as social media, would introduce new perspectives for the analysis.

**Ethical use of data and informed consent.** The intent of this research study is mainly meant to support decision-making during NFT trading, not to replace the human specialists. The authors are not responsible for any issues related to trading failures based on the results of this work.

**Table 5: Main characteristics of the NFT-set similarity graph by varying node size $l$: average degree (Deg), density (Den), diameter (D), average path length (APL), clustering coefficient (CC), modularity (M), no. of communities (#C)**

| $\mathcal{G}$ | $|V|$ | $|E|$ | Deg | Den | D | APL | CC | M | #C |
|---|---|---|---|---|---|---|---|---|---|
| $l = 100$ | 1808 | 13 939 | 15.419 | 0.009 | 42 | 12.756 | 0.457 | 0.795 | 11 |
| $l = 50$ | 3616 | 29 959 | 16.570 | 0.005 | 33 | 9.572 | 0.400 | 0.777 | 10 |
| $l = 25$ | 7232 | 62 322 | 17.235 | 0.002 | 27 | 8.051 | 0.356 | 0.768 | 11 |

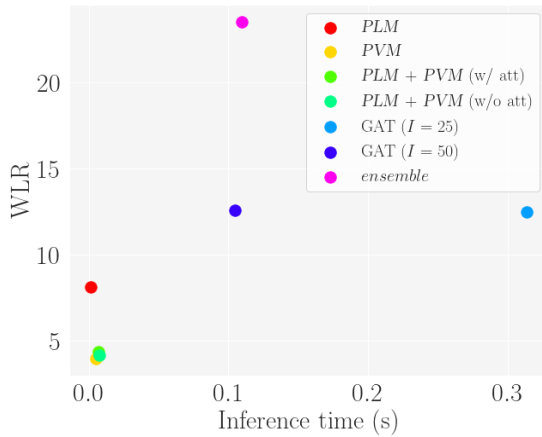

**Figure 4: Inference times w.r.t. $WLR$ scores for various models**

# APPENDIX

## A RUNNING TIMES AT INFERENCE

Figure 4 shows the running times at inference of various MERLIN models compared to their obtained $WLR$ performance. It stands out that, while the non-GAT modules are the most efficient ones, the GAT module leads to a significant improvement in $WLR$ at only a moderate inference-time cost (as long as the node size is not below the default of 50) w.r.t. the other modules. Also, the ensemble of the models' predictions allows for boosting the $WLR$ performance by doubling the GAT's one, and without any execution overhead thanks to parallelization of the visual and text modules.

## B ADDITIONAL REMARKS ON EVALUATION

**NFT-set similarity graph.** We additionally evaluated the effect on the structural traits of the resulting NFT-set similarity graph by different values for the node size $l$ (Table 5). One remarkable fact is that the number of communities, computed by the Louvain method [1], is roughly constant (set around 10) regardless of the choices for $l$ (note that the number of communities is not an input to the Louvain method). By delving into the community structures, Figure 5 (a-c) unveils that on each graph, with a more marked tendency for lower $l$, the discovered ten or eleven communities mostly capture the ten percentiles of the average selling price for the NFTs in our dataset. This further supports our initial intuition that there is an important impact of the visual and textual features on the selling price, i.e., *similar NFTs might sell similarly*. Moreover, we observed that the communities appear to be topologically contiguous according to the ordered percentiles: intuitively, this will have a positive impact on the effectiveness of the neighbor aggregation steps performed by the GNN module in MERLIN.

Interesting aspects also emerge about the graph that can be built from the similarities between the embeddings $\mathbf{h}^{(G)}$, i.e., by applying the GAT at inference. Indeed, Figure 5 (d) shows almost perfect separation of the NFT-sets w.r.t. their source collections, which is information never provided during the training phase. This is an outstanding evidence of the effectiveness of MERLIN to model contextual awareness in latent NFT representations, thus supporting our research hypotheses.

**Structural characteristics of NFT-set similarity graphs.** As reported in Table 5, we notice some interesting traits that would shed light on patterns underlying the NFT-group similarities. For instance, quite large values for diameter and average path length suggest that less similar nodes in our graphs tend to be relatively far apart. Moreover, good indications on triadic closure and modular mesoscopic structure arise from the clustering coefficient and modularity.

**Early-stopping.** We compared the effects of our defined early-stopping criterion based on win-rate w.r.t. the one based on accuracy. We find that (results not shown) the win-rate-based criterion is clearly more beneficial in terms of *High*-driven performance measures, with an improvement up to 48.7% of $WLR$, and on average 22.7% of riskiness and 9.2% of cautiousness over all models.

## C INTERPRETATIONS OF MERLIN PREDICTIONS

Here we discuss the interpretation of MERLIN predictions and their meaningfulness w.r.t. the task at hand, by resorting to *LIME* (Local Interpretable Model-Agnostic Explanations) [16]. Being model-agnostic, *LIME* just requires the class probabilities outputted by the model that is to be interpreted. It learns a linear model that approximates the target one in the neighborhood of the instance that needs to be explained, by perturbing the latter in order to learn feature importance scores w.r.t. the outcome of the model.

In this respect, in Figure 6 we present different examples of NFTs in our test set, and show the importance of features of images and descriptions w.r.t. class *High*, denoting with the green, resp. red, features that positively, resp. negatively, contributed to the class prediction performed by our best PVM and PLM, respectively.

Figure 6 (a) shows an NFT of the *Gods-Unchained* collection ($ID = 9493839$), i.e., a collectible card game, correctly classified by both PVM and PLM. Among the visual features positively impacting on the prediction capabilities, we report (i) the number 7 on the top-left side, which indicates the "mana cost" of a card, i.e., the amount of tokens needed for playing with it; (ii) the top-right symbol indicating the "set" of the card, i.e., its grouping w.r.t. particular events or themes; (iii) the wings of the portrayed creature and its name, along with the descriptive part of the character. Conversely, we spotted that the amount of damage tolerable from a creature (bottom-left) and its health score (bottom-right) may negatively

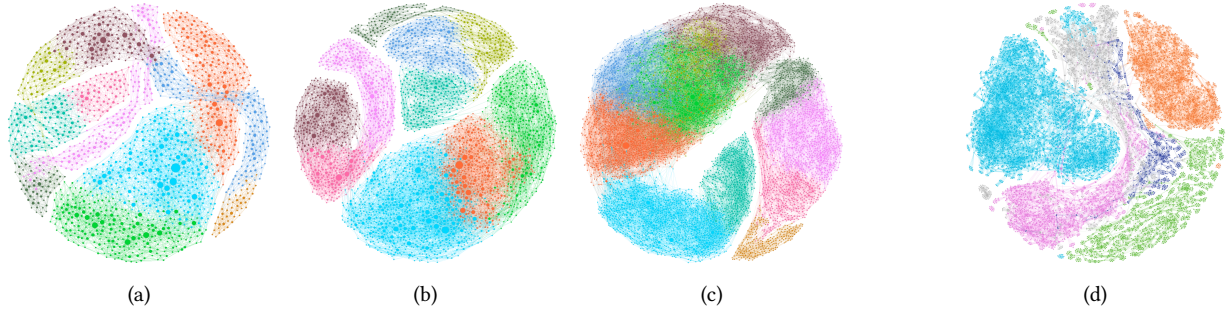

Figure 5: NFT-set similarity graphs, with colors corresponding to communities detected by Louvain method (a-c) and NFT-set similarity graph on validation set, with colors corresponding to top-5 collections (d). Plots (a-c) refer to different values for node size $l$: 100 (a), 50 (b), 25 (c), whereas plot (d) refers to $l = 50$

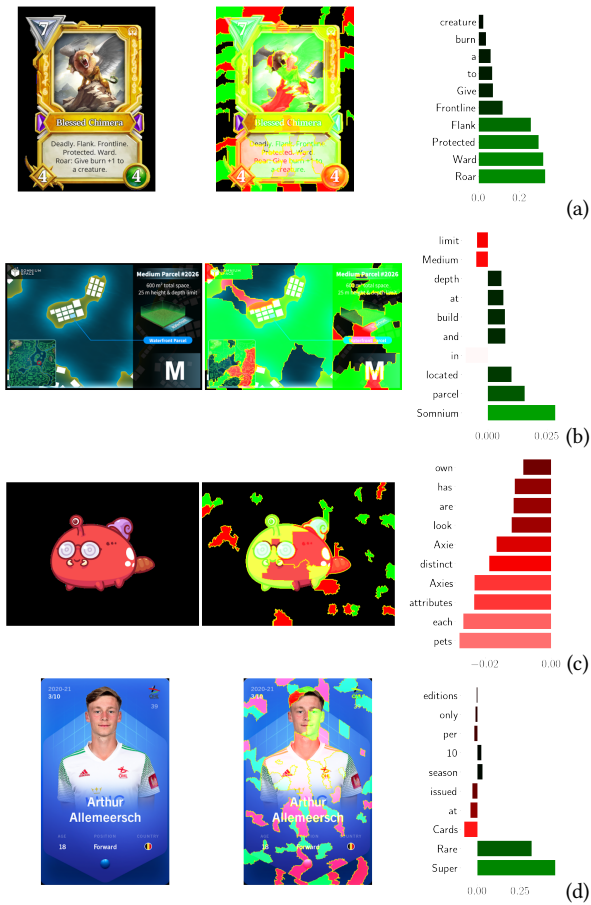

Figure 6: Example NFT images (left) and LIME interpretation of PVM features (center) and PLM features (right). Red, resp. green, denote a negative, resp. positive, impact of features on the model prediction

impact on the price prediction. As concerns the textual components, we report high importance for the terms "Roar", "Ward", "Protected" and "Flank", i.e., those describing cards' effects.

Figure 6 (b) shows an NFT of the *Somnium-Space* collection ($ID$ = 2026), i.e., a Metaverse project, correctly classified by both visual and textual modules in MERLIN. Here we spotted that the PVM actually benefits from the meaningful writings in the image (e.g., the square footage of the area), along with its shape and positioning in the mini-map. Conversely, the description seems to be less informative than the previous case (as it focuses mainly on the name of the Metaverse project), with scores produced by *LIME* for the PLM that are lower than an order of magnitude w.r.t. the previous example.

Looking at the example in Figure 6 (c), we observe an NFT whereby only PVM correctly predicted the target class. The *LIME* interpretation indeed provides some clues on that: the PVM seems to concentrate on the eyes, the silhouette and other characteristics of the profile of the drawing typical of the *Axie* collection; conversely, the description is poorly informative since it is the same across all NFTs in that collection, thus having a negative impact on the PLM performance.

Finally, in Figure 6 (d) we show an NFT of the *Sorare* collection, a football game in which players use NFTs to trade and manage virtual teams. Both PVM and PLM correctly predicted the target class. The *LIME* interpretation suggests that the PVM mostly focuses on the age and face of the player, as well as on his position on the pitch and team, as reported in the card. Besides, the "Super" and "Rare" terms are successfully leveraged by the PLM to discern that such an NFT is particularly rare within its collection.